\documentclass[conference]{IEEEtran}
\IEEEoverridecommandlockouts
\usepackage{amsmath,amssymb,amsfonts}
\usepackage{algorithm}
\usepackage{caption}
\usepackage{subfigure}
\usepackage{textcomp}
\usepackage{graphicx}
\usepackage{xcolor}
\usepackage{latexsym}
\usepackage{booktabs}
\usepackage{cite}
\usepackage{multirow}
\usepackage{algorithmicx}
\usepackage{algpseudocode}
\usepackage{amsthm}
\usepackage[justification=centering]{caption}

\def\BibTeX{{\rm B\kern-.05em{\sc i\kern-.025em b}\kern-.08em
    T\kern-.1667em\lower.7ex\hbox{E}\kern-.125emX}}
\begin{document}

\title{FairDD: Enhancing Fairness with domain-incremental learning in dermatological disease diagnosis\\
% {\footnotesize \textsuperscript{*}Note: Sub-titles are not captured in Xplore and
% should not be used}
% \thanks{Identify applicable funding agency here. If none, delete this.}
\thanks{%
        Tianlong Gu is the corresponding author.\\%
        This work was supported by the National Natural Science Foundation of China (Grant No.U22A2099, 62336003).
        }
}

\author{\IEEEauthorblockN{1\textsuperscript{st} Yiqin Luo}
\IEEEauthorblockA{\textit{College of Information Science and Technology}\\
\textit{Engineering Research Center of}\\ \textit{ Trustworthy AI (Ministry of Education)} \\
Guangzhou, China \\
lyq\_ah@163.com}
\and
\IEEEauthorblockN{2\textsuperscript{nd} Tianlong Gu}
\IEEEauthorblockA{\textit{College of Cyber Security}\\
\textit{Engineering Research Center of}\\ \textit{ Trustworthy AI (Ministry of Education)} \\
Guangzhou, China  \\
gutianlong@jnu.edu.cn}

% \author{\IEEEauthorblockN{1\textsuperscript{st} Yiqin Luo}
% \IEEEauthorblockA{\textit{College of Information Science and Technology} \\
% \textit{Jinan University}\\
% Guangzhou, China \\
% lyq\_ah@163.com}
% \and
% \IEEEauthorblockN{2\textsuperscript{nd} Tianlong Gu}
% \IEEEauthorblockA{\textit{College of Information Science and Technology} \\
% \textit{Jinan University}\\
% Guangzhou, China  \\
% gutianlong@jnu.edu.cn}
% \and
% \IEEEauthorblockN{3\textsuperscript{rd} Given Name Surname}
% \IEEEauthorblockA{\textit{dept. name of organization (of Aff.)} \\
% \textit{name of organization (of Aff.)}\\
% City, Country \\
% email address or ORCID}
% \and
% \IEEEauthorblockN{4\textsuperscript{th} Given Name Surname}
% \IEEEauthorblockA{\textit{dept. name of organization (of Aff.)} \\
% \textit{name of organization (of Aff.)}\\
% City, Country \\
% email address or ORCID}
% \and
% \IEEEauthorblockN{5\textsuperscript{th} Given Name Surname}
% \IEEEauthorblockA{\textit{dept. name of organization (of Aff.)} \\
% \textit{name of organization (of Aff.)}\\
% City, Country \\
% email address or ORCID}
% \and
% \IEEEauthorblockN{6\textsuperscript{th} Given Name Surname}
% \IEEEauthorblockA{\textit{dept. name of organization (of Aff.)} \\
% \textit{name of organization (of Aff.)}\\
% City, Country \\
% email address or ORCID}
}

\maketitle

\begin{abstract}
With the rapid advancement of deep learning technologies, artificial intelligence has become increasingly prevalent in the research and application of dermatological disease diagnosis. However, this data-driven approach often faces issues related to decision bias. Existing fairness enhancement techniques typically come at a substantial cost to accuracy. This study aims to achieve a better trade-off between accuracy and fairness in dermatological diagnostic models. To this end, we propose a novel fair dermatological diagnosis network, named FairDD, which leverages domain incremental learning to balance the learning of different groups by being sensitive to changes in data distribution. Additionally, we incorporate the mixup data augmentation technique and supervised contrastive learning to enhance the network’s robustness and generalization. Experimental validation on two dermatological datasets demonstrates that our proposed method excels in both fairness criteria and the trade-off between fairness and performance.
\end{abstract}


\begin{figure*}[t]
\centering
\includegraphics[width=0.8\textwidth]{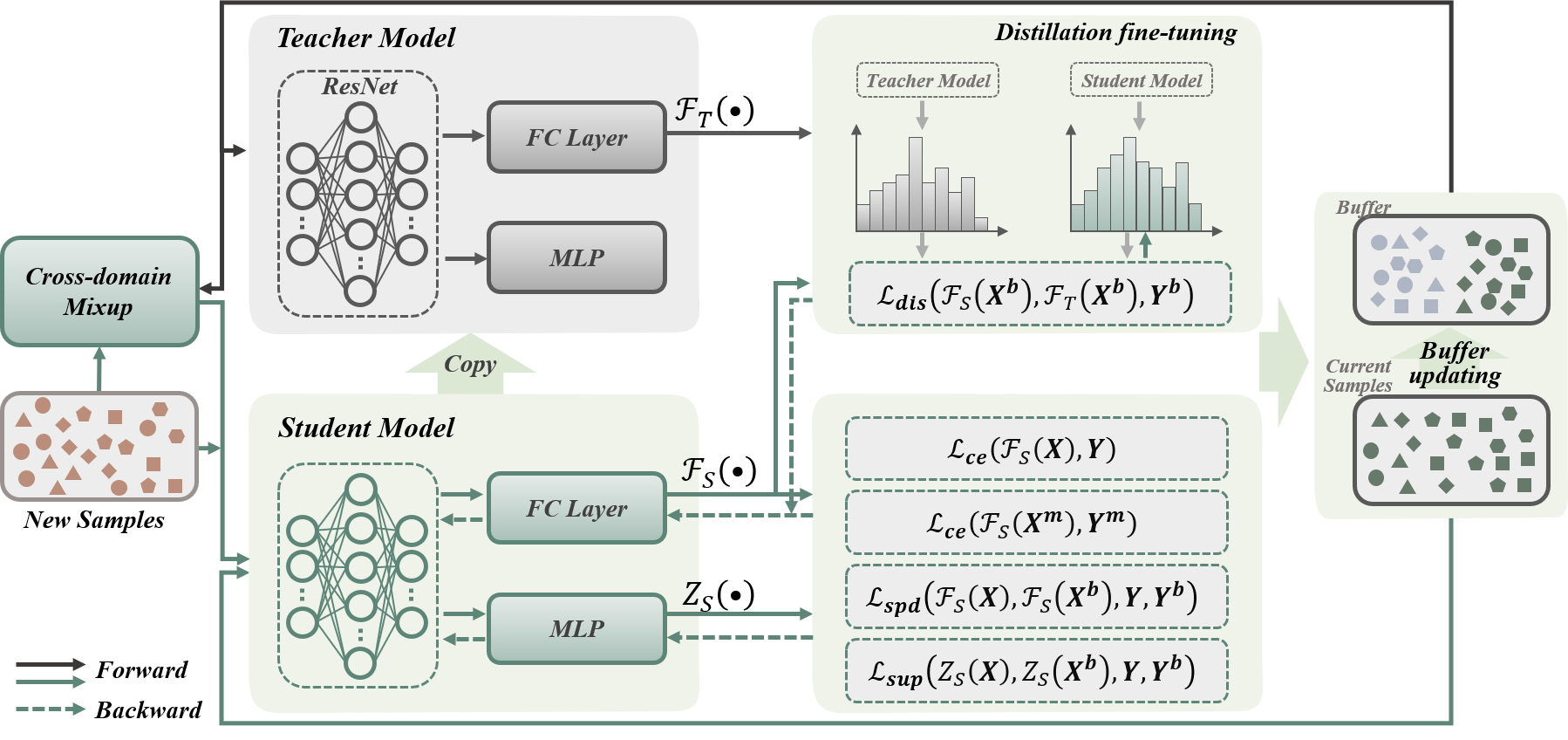}
\caption{The main framework of Network. At each stage, the student model is jointly trained on the input new domain samples and old domain samples from memory. During training, the teacher model provides knowledge to the student model while keeping its parameters fixed. After each stage of training, the student model’s parameters are copied to serve as the teacher model for the next stage. At the end of each epoch within a stage, the model undergoes distillation fine-tuning, and the memory is updated. Note that memory updates cease in the final phase. In the first stage, the student model is trained independently.}
\label{fig:1}
\end{figure*}

\section{Introduction}
Dermatological diseases are a major public health issue, threatening the health of millions globally. With the increasing performance refinement of deep learning techniques, artificial intelligence has achieved significant results in aiding the diagnosis of dermatological disease \cite{BA1}. This data-driven approach enables networks to learn task-specific features, achieving high accuracy in target tasks. However, when applied to large datasets, these methods exhibit a strong dependency on data quality and may suffer from biases due to the imbalanced distribution of training data \cite{B6}. A critical concern is that, to enhance model accuracy, neural networks may rely on demographic attributes, particularly sensitive features like skin color or gender, to differentiate between diseases \cite{BA3}. This reliance can lead to notable performance disparities across different demographic attribute subgroups, i.e., model bias. When such biased models are deployed in real-world, they can have adverse effects on individuals and society. Therefore, there is an urgent need for an in-depth study of diagnostic bias.

Numerous studies in the academic domain have focused on diagnostic model fairness \cite{BA6}. For instance, Wu et al. \cite{BC6} proposed a method to prune pre-trained diagnostic models by assessing feature importance disparities between subgroups, thereby mitigating model unfairness. However, as a post-processing method \cite{BA4}, this approach inevitably increases extra computational time. In contrast, FairAdaBN \cite{B1} employs an in-processing method that dynamically integrates common information across subgroups by sharing network parameters while introducing only a few additional parameters for subgroup-specific feature map representation. 
Although these methods demonstrate significant improvements in fairness, they often incur substantial accuracy losses. Thus, achieving a balance between fairness enhancement and accuracy retention remains challenging \cite{B1}. Our research aims to maximize fairness while preserving model performance, striving for a more optimal trade-off between accuracy and fairness.

Firstly, we focus on the fairness problem. The core principle of most fairness enhancement methods involves learning from and adapting to the inherent imbalances in data distribution. This is typically achieved by either balancing the data distribution or adapting the learning algorithm \cite{B2,BC7}.
Similarly, domain incremental learning follows this thought, enabling continuous adaptation to new information and integration of new knowledge without compromising previously acquired information, thereby achieving balanced learning across different domains \cite{BB1}. Thus, we propose leveraging the unique property of domain incremental learning to mitigate bias. 

However, incremental learning faces a significant challenge: ``catastrophic forgetting," \cite{BA5} where acquiring new knowledge often leads to the gradual erosion of previously learned information, potentially harming the model's overall performance. To address this issue, we integrate a memory replay \cite{BB6} with the Mixup data augmentation technique to enhance the model's anti-forgetting ability and robustness. Furthermore, to counter the overfitting commonly seen in deep learning models, we introduce contrastive loss to bolster the model's generalization capability. Finally, by incorporating a distillation fine-tuning module, we effectively balance the model's performance. Our main contributions are as follows:

\begin{itemize}

\item we propose a fair domain-incremental-based dermatological diagnosis model, named FairDD, which aims to achieve a better trade-off between accuracy and fairness.
\item A cross-domain Mixup module is introduced, facilitating the model in learning more robust feature representations by mixing data from both new and old domains.
\item We designed a distillation fine-tuning module, which balances the weights of old and new knowledge by distilling old knowledge and fine-tuning new knowledge, thus enabling the model to learn new information while retaining previously acquired knowledge.
\item Experiments on two dermatological datasets demonstrate that FairDD exhibits significant advantages in the trade-off between diagnostic performance and fairness.
\end{itemize}

\section{Methods}
\subsection{Model Overview}
In domain incremental learning, data is loaded incrementally in batches. Each batch represents a new domain or a set of new domains. After completing the training on one batch of data, the model will enter the next training stage. The current batch data is considered a new domain, while previously learned data is collectively termed old domains. 
Additionally, the data domains of the training and test sets are updated simultaneously to ensure accurate evaluation.

Given a dermatological image dataset $S$ containing $N$ samples, each sample includes an input image $X$, a sensitive attribute $A$, and a categorized truth label $Y$. The dataset $S$ is systematically divided into different domain groups based on the sensitive attribute $A$. For instance, based on gender, the data is classified into male and female domains. This study primarily aims to achieve an optimal trade-off between the model’s diagnostic performance and fairness. Specifically, we seek to maintain the overall performance of the model while minimizing differences in diagnostic accuracy across various sensitive attribute groups.

Our FairDD incorporates a bounded replay buffer $M=\{( x_i^b,a_i^b,y_i^b )_{i=1}^B\}$ of fixed size $B$ to prevent catastrophic forgetting. The buffer is updated using a reservoir algorithm that randomly selects samples from the training data stream with equal probability. The main framework is shown in Fig. \ref{fig:1}.

\subsection{Cross-Domain Mixup Module}
Data augmentation extends a training dataset by generating new examples using prior knowledge. Mixup, a specific data augmentation method, creates new training samples through convex combinations of sample pairs \cite{new1}. 
Specifically, given two training samples and their corresponding labels, mixup constructs virtual training samples by linear interpolation. 
This strategy helps to improve the model’s generalization ability.

In this paper's setup, the model incrementally learns data from different domains. During training with new domain data, only a small portion of old domain data is retained in memory and participates in the training. This can lead the model to favor new domain data. To address this issue, we introduce the mixup method, which combines data from both new and old domains to encourage the model to learn a more robust feature representation. Additionally, the interpolated samples generated by mixup enhance the diversity of the training data and help mitigate the risk of overfitting to the original data. The implementation details of mixup are provided below:
\begin{equation}
\left\{\begin{array}{l}
x^m=\lambda x_i+(1-\lambda) x_j \\
y^m=\lambda y_i+(1-\lambda) y_j
\end{array}, \lambda \sim \operatorname{Beta}(\theta, \theta)\right. 
\label{eq:1}
\end{equation}
where $(x_i,a_i,y_i )\in XB$, is the input data of the current stage. $(x_j,a_j,y_j )\in XB\cup M$. $\lambda\in[0,1]$ and $\theta$ is set to 0.8.

\subsection{Supervised Contrastive Loss}
Supervised contrastive learning leverages label information to enhance model representation learning \cite{new2}. 
% It treats all data sharing the same label as positive pairs, calculates their similarities, and performs a weighted average. 
This method pulls together points within the same category in the embedding space while pushing apart clusters of samples from different categories, enabling the model to learn more compact and discriminative data representations.

Building on this concept, we incorporate a contrastive loss into the model.
% to enhance the student model's ability to cluster embeddings of the same class while distancing those from different classes. 
Specifically, we introduce a learnable projector that maps sample features to the embedding space where the contrastive loss is applied, thereby learning marginal distributions for improved generalization. The projectors are updated using gradient-based optimization. The supervised contrastive loss is calculated as follows:
\begin{equation}
\mathcal{L}_{\text {sup }}=\sum_{i \in I} \frac{-1}{|P(i)|} \sum_{p \in P(i)} \log \frac{\exp \left(\operatorname{sim}\left(z_i, z_p\right) / \tau\right)}{\sum_{o \in O(i)} \exp \left(\operatorname{sim}\left(z_i, z_o\right) / \tau\right)}  
\label{eq:2}
\end{equation}

where $I$ denotes the index set of the current batch and the memory buffer batch data. $P(i)$ represents the indexed set of samples sharing the same label (i.e., positive samples) as sample $i$, while $O(i)=P(i) \cup\{j \mid j \in I \wedge j \neq i\}$ includes the indexed set of all samples (both positive and others) in the same batch as sample $i$. $z$ denotes the feature representation of the sample obtained through the projector. The function $sim()$ measures cosine similarity, and $\tau$ is a temperature hyperparameter, set to 0.07, to control the concentration degree of the distribution.

\subsection{Statistical Parity Disparity Loss}\label{SCM}
A core objective of this paper is to enhance the fairness of the student model, ensuring equitable treatment of samples across different sensitive attribute domains. Specifically, we focus on statistical parity disparity, a definition that measures the difference in predictions across sensitive attribute domains when the model predicts a positive (or target) class \cite{A6}.

Therefore, we introduce statistical parity disparity loss as a fairness optimization objective to minimize the discrepancy between the model's predictions for target classes across different sensitive attribute groups. The input data here includes not only data from the current domain but also data from the buffer M. The statistical parity disparity loss is:
\begin{eqnarray}\label{eq:3}  
\begin{split}   
& \mathcal{L}_{s p d}=\\
& \sum_{y=1}^U\left\|\mathbb{E}_{X_i \sim S_{A=0}}\left|\left(q\left(X_i\right)=y\right)-\mathbb{E}_{X_i \sim S_{A=1}}\right|\left(q\left(X_i\right)=y\right)\right\|^2  
\end{split}
\end{eqnarray}

\begin{table*}[t]
\caption{Result on Fitzpatrick-17k and ISIC 2019 Dataset ($Mean^{Std}\times10^{-2}$).\ \textbf{Best}\ and \ \underline{Second-best} \ are \ highlighted.\ E0: $FATE_{EOpp0}$.\ E1: $FATE_{EOpp1}$.\ $E2: FATE_{EOdd}$.}
\label{tab:1}
\centering
\begin{tabular}{c|cccc|ccc|ccc}
\toprule
\multicolumn{11}{c}{\textbf{Fitzpatrick-17k Dataset} } \\ \hline
Method                  & Accuracy $\uparrow$ & Precision $\uparrow$ & Recall $\uparrow$  & F1 $\uparrow$      & EOpp0 $\downarrow$  & EOpp1 $\downarrow$   & Eodd $\downarrow$    & E0 $\uparrow$   & E1 $\uparrow$  & E2 $\uparrow$  \\ \hline
Vanilla                 & $87.53^{0.14}$  & $\textbf{79.60}^{\textbf{0.33}}$   & $\textbf{80.22}^{\textbf{0.19}}$ & $\textbf{78.41}^{\textbf{0.15}}$ & $1.00^{0.30}$ & $10.40^{1.43}$ & $10.54^{0.98}$ & /       & /      & /      \\
Resampling\cite{B2}      & \underline{$87.73^{0.27}$}  & \underline{$79.21^{0.40}$}   & \underline{$80.01^{0.35}$} & \underline{$78.27^{0.42}$} & $1.11^{0.26}$ & $10.43^{1.91}$ & $10.78^{2.06}$ & $-10.86$  & $-0.03$  & $-2.05$  \\
Ind\cite{B2}             & $86.33^{0.12}$  & $76.11^{0.38}$   & $77.48^{0.18}$ & $75.20^{0.09}$ & $0.78^{0.33}$ & $10.13^{0.51}$ & $9.72^{0.94}$  & 20.63   & 1.23   & 6.41   \\
GroupDRO\cite{B3}        & $86.62^{0.19}$  & $77.21^{0.62}$   & $78.29^{0.52}$ & $76.56^{0.56}$ & $0.94^{0.34}$ & $8.04^{0.90}$  & $8.23^{1.25}$ & 5.07    & 21.66  & 20.91  \\
EnD\cite{B4}             & $86.80^{0.52}$  & $77.32^{0.60}$   & $78.58^{0.53}$ & $76.90^{0.66}$ & $1.22^{0.31}$ & $9.01^{1.60}$  & $9.20^{1.59}$  & -22.83  & 12.53  & 11.88  \\
CFair\cite{B5}           & $\textbf{87.91}^{\textbf{0.35}}$  & $78.62^{0.49}$   & $79.73^{0.37}$ & $78.12^{0.38}$ & $0.93^{0.28}$ & $9.83^{1.65}$  & $10.17^{1.57}$ & 10.03   & 12.15  & 10.09  \\
FairAdaBN\cite{B1}       & $84.72^{0.40}$  & $74.43^{0.22}$   & $75.74^{0.33}$ & $73.31^{0.48}$ & \underline{$0.48^{0.09}$} & \underline{$7.67^{3.86}$}  & \underline{$7.73^{3.95}$}  & \underline{48.79}   & \underline{23.04}  & \underline{23.45}  \\
QP-Net\cite{B6}                  & $83.16^{0.06}$  & $70.17^{0.10}$   & $72.61^{0.20}$ & $70.41^{0.18}$ & $0.61^{0.08}$ & $9.11^{0.23}$  & $9.41^{0.33}$  & 34.01   & 7.41   & 5.73   \\
FairDD                 & $86.53^{0.16}$  & $76.70^{0.14}$   & $77.16^{0.18}$ & $76.75^{0.06}$ & $\textbf{0.48}^{\textbf{0.07}}$ & $\textbf{5.73}^{\textbf{1.43}}$  & $\textbf{5.65}^{\textbf{1.43}}$  & \textbf{51.19}   & \textbf{43.72}  & \textbf{45.22}  \\ \hline
\multicolumn{11}{c}{\textbf{ISIC 2019 Dataset}} \\ \hline
Method                  & Accuracy $\uparrow$ & Precision $\uparrow$ & Recall $\uparrow$  & F1 $\uparrow$      & EOpp0 $\downarrow$  & EOpp1 $\downarrow$   & Eodd $\downarrow$    & E0 $\uparrow$   & E1 $\uparrow$  & E2 $\uparrow$  \\ \hline
Vanilla                 & \underline{$92.52^{0.12}$}  & \underline{$82.64^{0.31}$}   & \underline{$82.94^{0.36}$} & \underline{$82.60^{0.32}$} & $0.85^{0.12}$ & $6.12^{1.83}$  & $6.02^{1.66}$  & /       & /      & /      \\
Resampling\cite{B2}      & $\textbf{92.81}^{\textbf{0.28}}$  & $\textbf{83.15}^{\textbf{0.50}}$   & $\textbf{83.42}^{\textbf{0.51}}$ & $\textbf{83.12}^{\textbf{0.52}}$ & $0.86^{0.15}$ & $5.65^{2.83}$  & $5.76^{2.78}$  & -0.80    & -2.48  & -5.49  \\
Ind\cite{B2}             & $92.43^{0.11}$  & $82.16^{0.15}$   & $82.46^{0.12}$ & $82.11^{0.08}$ & $0.85^{0.11}$ & $7.04^{0.96}$  & $7.37^{0.77}$  & -0.10   & -15.13 & -22.52 \\
GroupDRO\cite{B3}        & $91.86^{0.22}$  & $81.30^{0.52}$   & $81.44^{0.47}$ & $81.17^{0.50}$ & $0.82^{0.12}$ & $6.78^{3.20}$  & $6.62^{3.21}$  & 2.41    & -22.99 & -22.01 \\
EnD\cite{B4}             & $92.13^{0.08}$  & $81.42^{0.48}$   & $81.64^{0.35}$ & $81.36^{0.38}$ & $0.98^{0.09}$ & $5.18^{0.99}$  & $5.10^{1.06}$  & -15.72  & 14.94  & 14.86  \\
CFair\cite{B5}           & $87.39^{0.77}$  & $72.39^{2.67}$   & $72.60^{2.22}$ & $71.28^{2.12}$ & $2.83^{1.09}$ & $9.21^{3.53}$  & $10.80^{4.15}$ & -238.49 & -56.03 & -84.95 \\
FairAdaBN\cite{B1}       & $89.11^{0.09}$  & $74.24^{0.13}$   & $74.79^{0.18}$ & $74.18^{0.14}$ & \underline{$0.69^{0.07}$} & $4.85^{2.50}$  & \underline{$4.76^{2.73}$}  & \underline{15.14}   & 17.07  & \underline{17.24}  \\
QP-Net \cite{B6}                  & $88.00^{0.22}$  & $74.05^{0.27}$   & $72.55^{0.26}$ & $71.49^{0.31}$ & $1.13^{0.10}$ & $\textbf{4.56}^{\textbf{0.54}}$  & $4.94^{0.38}$  & -37.83  & \underline{20.60}   & 13.05  \\
FairDD                 & $90.70^{0.21}$  & $81.28^{0.49}$   & $80.70^{0.40}$ & $80.84^{0.41}$ & $\textbf{0.59}^{\textbf{0.04}}$ & $\underline{4.58^{0.25}}$  & $\textbf{4.21}^{\textbf{0.16}}$  & \textbf{28.62}   & \textbf{23.25}  & \textbf{28.10} \\
\bottomrule
\end{tabular}
\end{table*}

\subsection{Distillation Fine-Tuning}
Given that only a limited number of samples from the old domain are stored in memory, there are significantly fewer old domain samples than new domain samples in the training set. To address this imbalance, a balancing fine-tuning phase is commonly introduced
\cite{A4}. 
% \cite{A3,A4,A5}. 
During this phase, the model is fine-tuned on a carefully constructed balanced subset containing an equal number of samples from each class of both the new and old domains, using a smaller learning rate to optimize performance.

In this paper, the parameters of the student model are initially aligned with those of the teacher model at the start of each new learning stage. As the student model learns from the new domain data, its parameters are iteratively updated. However, the training samples from the old and new domains are often unbalanced during this process. Therefore, it is crucial to ensure that the student model retains the knowledge of the old domain while incorporating new domain knowledge.

In general, samples from the same class should have similar embeddings in the feature space. Therefore, the student model's embeddings should align with those of the same class in the teacher model. Distillation loss effectively measures the difference in output distributions between the student and teacher models, aiding the student model in approximating the teacher model's output. Building on the benefits of balanced fine-tuning, we introduce a distillation fine-tuning component that directly uses small batches of buffer data as inputs without requiring additional balanced subsets. The specific distillation loss function is defined as follows:
\begin{eqnarray}\label{eq:4}
\mathcal{L}_{\text {dis }}=-\sum_{i=1}^N \sum_{j=1}^U q_t\left(x_{i, j}^{\prime}\right) \log \left(q_s\left(x_{i, j}^{\prime}\right)\right)
\end{eqnarray}
\begin{eqnarray}\label{eq:5}
q_t\left(x_{i, j}^{\prime}\right)=\frac{\left(q_t\left(x_{i, j}\right)\right)^{\frac{1}{T}}}{\sum_i\left(q_t\left(x_{i, j}\right)\right)^{\frac{1}{T}}},
q_s\left(x_{i, j}^{\prime}\right)=\frac{\left(q_s\left(x_{i, j}\right)\right)^{\frac{1}{T}}}{\sum_i\left(q_s\left(x_{i, j}\right)\right)^{\frac{1}{T}}}
\end{eqnarray}
where $T$ represents the distillation temperature coefficient (set to 2). $q_{t} (x_{i,j} )$ and $q_{s} (x_{i,j} )$ represent the predicted probabilities of the teacher and the student model, respectively.

\subsection{Overall Objective}
In summary, the overall training objective of our proposed FairDD can be expressed as:
\begin{eqnarray}\label{eq:7}
\mathcal{L}=\mathcal{L}_{c e}+ \mathcal{L}_{sup}+\alpha \mathcal{L}_{d i s}+\beta \mathcal{L}_{s p d}
\end{eqnarray}
\begin{eqnarray}\label{eq:8}
\mathcal{L}_{c e}=-\frac{1}{N} \sum_{i=1}^N \sum_{j=1}^U p\left(x_{i, j}\right) \log \left(q\left(x_{i, j}\right)\right)
\end{eqnarray}
where weighting factors $\alpha$ and $\beta$ adjust the degree of constraints on knowledge transfer and fairness. $p\left(x_{i, j}\right)$ and $q\left(x_{i, j}\right)$ denote the true and predicted probabilities respectively.

\section{Experiments and Results}
We evaluated our proposed FairDD on the datasets Fitzpatrick-17k \cite{D1} and ISIC 2019 
\cite{D3}.
% \cite{D2, D3, D4}. 
The sensitive attributes of these two datasets are skin type and age,
respectively. The parameter $\alpha$ and cache size were set to 1 and 300, respectively. For the Fitzpatrick-17k and ISIC 2019 datasets, the parameter $\beta$ was set to 0.6 and 1, respectively.

\subsection{Evaluation Metrics}
To provide a comprehensive assessment of model performance, we conducted an in-depth analysis from multiple perspectives, including diagnostic accuracy, fairness, and trade-off. Accuracy, precision, recall, and F1 scores were used in evaluating diagnostic performance.
The fairness criteria include Equal Opportunity (EOpp) \cite{M1} and Equal Odds (EOdd) \cite{M1}. For Equal Opportunity, we consider both EOpp0 and EOpp1 based on actual labels.
The indicator $FATE$ \cite{B1} is used to quantify the trade-off.

The formula is as follows:
\begin{eqnarray}\label{eq:16}
F A T E_{F C}=\frac{A C C_e-A C C_b}{A C C_b}-\lambda \frac{F C_e-F C_b}{F C_b}
\end{eqnarray}
where $FC$ denotes fairness criteria. $ACC$ indicates accuracy. Subscripts $e$ and $b$ are used to distinguish the fairness-enhanced model from the baseline model. $\lambda$ is set to 1.0.

\begin{table*}[t]
\centering
\caption{Impact of training order, cross-domain mixup module and Supervised Contrastive Loss.}
\label{tab:2}
\begin{tabular}{c|c|c|cccc|ccc|ccc}
\hline
order                       & Mixup & $\mathcal{L}_{sup}$ & Accuracy $\uparrow$ & Precision $\uparrow$ & Recall $\uparrow$ & F1 $\uparrow$   & EOpp0 $\downarrow$ & EOpp1 $\downarrow$ & Eodd $\downarrow$ & E0 $\uparrow$    & E1 $\uparrow$    & E2 $\uparrow$    \\ \hline
\multirow{3}{*}{light-dark} &       &       & 83.69     & 73.41      & 72.95   & 73.08 & 1.00   & 10.87  & 11.09 & $-4.38$  & $-8.90$  & $-9.60$  \\  
                            & $\checkmark$     &       & 84.88     & 74.81      & 74.98   & 74.86 & 1.24   & 12.08  & 13.08 & $-27.02$ & $-19.18$ & $-27.12$ \\ 
                            & $\checkmark$     & $\checkmark$     & 85.41     & 75.23      & 76.30    & 75.29 & 1.47   & 10.98  & 12.14 & $-49.42$ & $-7.99$  & $-17.60$ \\ \hline
\multirow{3}{*}{dark-light} &       &       & 85.48     & 75.06      & 75.39   & 75.16 & 0.69   & 6.06   & 6.58  & 28.65  & 39.38  & 35.22  \\  
                            & $\checkmark$     &       & 86.10      & 76.36      & 76.68   & 76.46 & 0.58   & 6.63   & 6.14  & 40.36  & 34.61  & 40.11  \\  
                            & $\checkmark$     & $\checkmark$     & 86.56     & 76.77      & 77.36   & 76.80  & 0.47   & 6.05   & 5.99  & 51.89  & 40.71  & 42.06  \\ \hline
\end{tabular}%
\end{table*}

\begin{figure*}[t]
\centering
\subfigure[Diagnostic accuracy of FairDD]{
\includegraphics[width=0.3\textwidth]{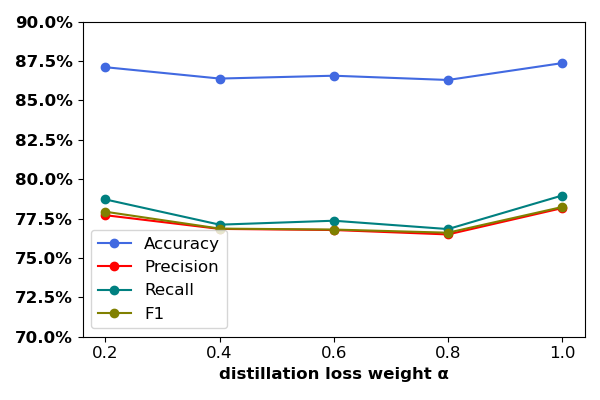}}
\subfigure[Fairness of FairDD]{
\includegraphics[width=0.3\textwidth]{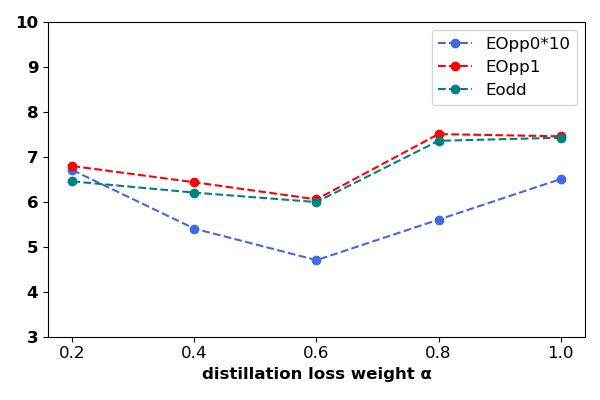}}
\subfigure[Trade-off of FairDD]{
\includegraphics[width=0.3\textwidth]{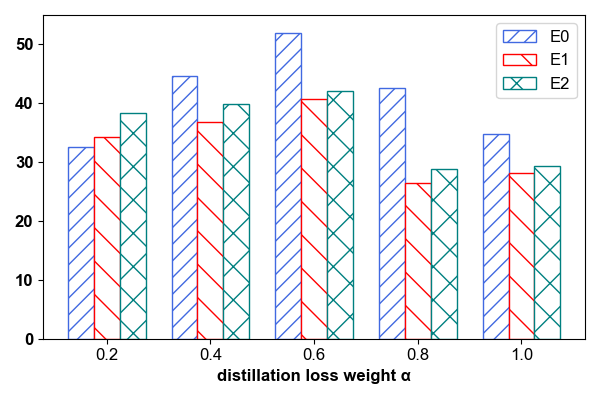}}
% \caption{Impact of distillation loss.}
\caption{Impact of distillation fine-tuning module.}
\label{fig:3}
\end{figure*}

\begin{figure*}[t]
\centering
\subfigure[Diagnostic accuracy of FairDD]{
\includegraphics[width=0.3\textwidth]{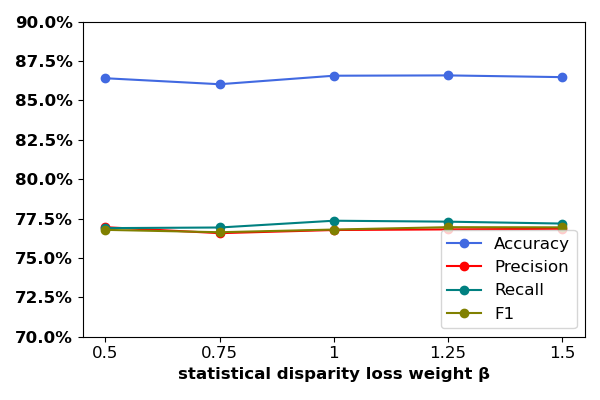}}
\subfigure[Fairness of FairDD]{
\includegraphics[width=0.3\textwidth]{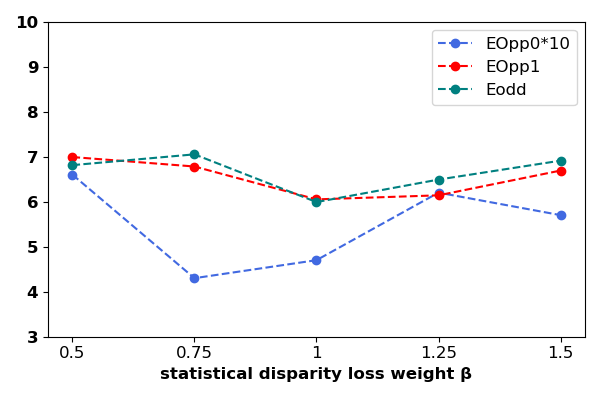}}
\subfigure[Trade-off of FairDD]{
\includegraphics[width=0.3\textwidth]{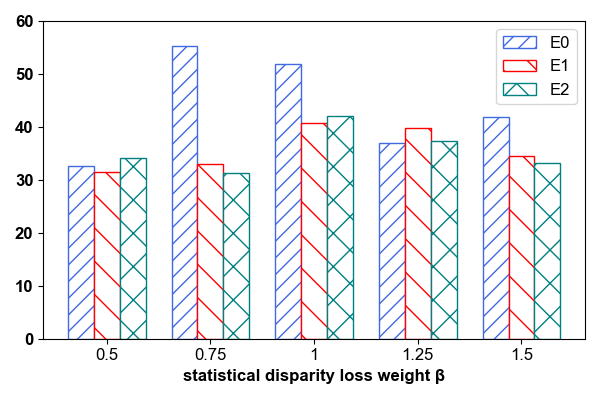}}
\caption{Impact of statistical disparity loss.}
\label{fig:4}
\end{figure*}

\begin{figure*}[t]
\centering
\subfigure[Diagnostic accuracy of FairDD]{
\includegraphics[width=0.3\textwidth]{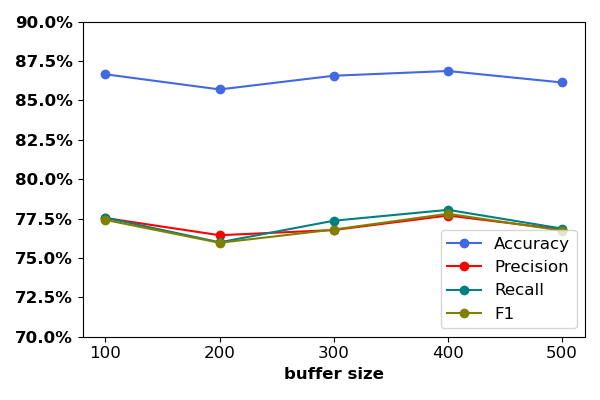}}
\subfigure[Fairness of FairDD]{
\includegraphics[width=0.3\textwidth]{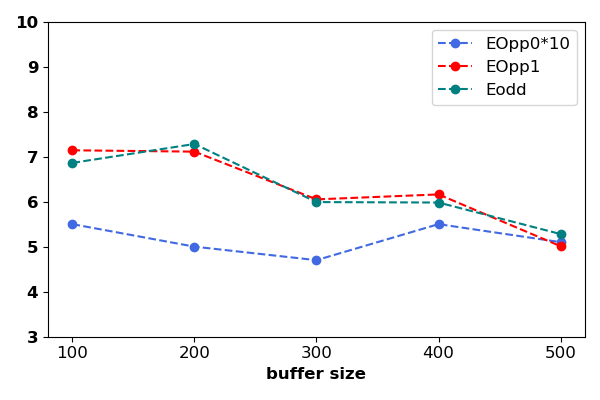}}
\subfigure[Trade-off of FairDD]{
\includegraphics[width=0.3\textwidth]{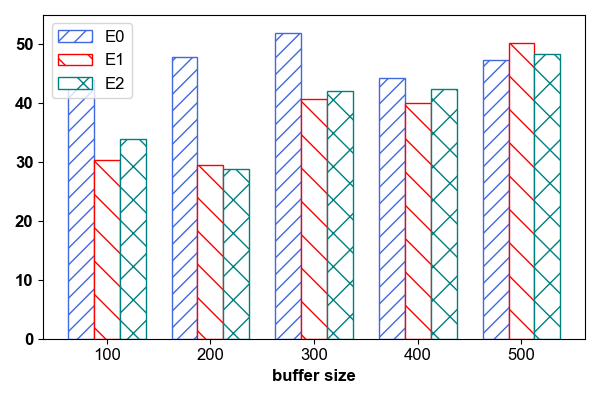}}
\caption{Impact of buffer size.}
\label{fig:5}
\end{figure*}

% \section{Results and Analysis}
\subsection{Comparison with Other Methods}

\textbf{Results on Fitzpatrick-17k Dataset.} 
FairDD demonstrates superior performance to other methods on all fairness assessment criteria, as shown in Table \ref{tab:1}. Specifically, compared with the suboptimal results, FairDD decreased by about 2\% in both EOpp1 and Eodd. Second, although FairDD does not achieve the best in terms of diagnostic performance, it improves on accuracy, precision, and F1 scores when compared to FairAdaBN, the model with sub-optimal fairness. There was a 3.4\% improvement in F1. Further analyzing the trade-off between diagnostic accuracy and fairness, we observe that FairDD achieves the highest values in each trade-off criterion. These combined results provide strong evidence that FairDD significantly improves the fairness of the model while maintaining diagnostic performance.

\textbf{Results on ISIC 2019 Dataset.} 
Unbalanced data distribution is a major factor affecting the fairness of decision making. Compared to the Fitzpatrick-17k dataset, the distribution of sensitive attributes of the ISIC 2019 Dataset is more balanced, but the comparative models perform poorly, as shown in  Tabl \ref{tab:1}. Against this challenging backdrop, FairDD still stands out as the best result for all fairness and performance trade-offs, except for EOpp1, which is 0.02 higher than the next best result. Although FairDD has not yet surpassed the Vanilla model in terms of accuracy, it has significantly improved its performance compared to the FairAdaBN and QP-Net models, which are the next best in terms of fairness. These demonstrate its advantage in the trade-off between diagnostic performance and fairness.

\subsection{Ablation Study}
\textbf{Impact of training order.} Incremental learning technology can balance the learning of each domain and effectively transfer previously learned feature knowledge to subsequent training. We found that the effect of incremental learning on balanced learning is closely related to the training order. Specifically, as demonstrated in Table \ref{tab:2}, the model's performance and its balance across different skin color groups are notably better when the training sequence follows a dark-light order compared to the opposite order. 

\textbf{Impact of cross-domain mixup module.} Compared to FairDD without mixup, the diagnostic performance of FairDD with mixup integration shows consistent improvement across all training orders. This finding indicates that the interpolated samples generated by mixup enhance the diversity of the training data and encourage the model to learn more robust feature representations.

\textbf{Impact of supervised contrastive loss.} Adding supervised contrastive loss improves the model's accuracy, fairness, and trade-off metrics, except for EOpp0 in the light-dark order. This enhancement indicates that supervised contrastive loss encourages embeddings of the same class to converge, leading to better generalization. However, the effectiveness of this loss may be influenced by the training order.

\textbf{Impact of distillation fine-tuning module.} 
In Equation 7, the coefficient $\alpha$ of the distillation loss controls the contribution of the balanced fine-tuning component to the overall performance. As $\alpha$ increases from 0.2 to 1, diagnostic performance metrics initially decrease, then slowly increase, and generally remain stable. At $\alpha$=0.6, the fairness metrics and trade-offs reach optimal values, as shown in Fig. 3b and 3c. However, as the value of $\alpha$ continues to increase, the model's fairness rapidly declines, despite a gradual improvement in diagnostic performance. This is because the balanced fine-tuning process relies on data from the old domain. Excessive fine-tuning stabilizes performance in the old domain but significantly diminishes performance in the new domain. 

\textbf{Impact of statistical parity disparity loss.} 
For this part of the experiment, the coefficient $\alpha$ is set to 0.6.
As illustrated in Fig. 4a, changes in the coefficient $\beta$ have a minimal impact on the model's diagnostic performance, which remains stable. Fairness initially improves and then diminishes. Thus, while the constraint imposed by statistical disparity loss enhances fairness, excessive constraint limits this enhancement. Considering diagnostic performance and fairness, the trade-off of model also first increases and then gradually declines, as depicted in Fig. 4c.

\textbf{Impact of buffer size.} 
In this part, the $\alpha$ and $\beta$ are set to 0.6 and 1. 
In Fig. 5a, the accuracy fluctuates within a specific range as buffer size gradually increases. Regarding fairness, EOpp1 and Eodd values decrease as buffer size increases, while EOpp0 initially decreases and then slowly increases once the capacity exceeds 300. Increased buffer size provides more samples from old attribute classes, enriching the information available for balance fine-tuning and statistical disparity loss. 
The model's trade-off metric rises initially and then stabilizes during this process.

\section{Conclusion}
In this paper, we propose a fair domain-incremental-based dermatological diagnosis model named FairDD, aiming to achieve a better trade-off between diagnostic performance and fairness. The model is optimized using various components to enhance both diagnostic accuracy and fairness. The effectiveness of each component is also validated by exhaustive ablation experiments. Experimental results on two dermatological disease datasets demonstrate that FairDD achieves favorable outcomes in fairness and trade-offs. However, there remains a gap in diagnostic performance between FairDD and the Vanilla model. Therefore, future research will focus on enhancing FairDD's diagnostic accuracy while maintaining fairness and trade-off.

% \section*{Acknowledgment}

% This work was supported in part by the National Natural Science Foundation of China under Grants U22A2099 and 62336003.

\bibliographystyle{IEEEtran}
\bibliography{FDD}

\end{document}